\documentclass[]{article}
\pdfoutput=1
\usepackage{lmodern}
\usepackage{amssymb,amsmath}
\usepackage{ifxetex,ifluatex}
\usepackage{fixltx2e} % provides \textsubscript

\usepackage[square,sort,comma,numbers]{natbib}

\ifnum 0\ifxetex 1\fi\ifluatex 1\fi=0 % if pdftex
  \usepackage[T1]{fontenc}
  \usepackage[utf8]{inputenc}
\else % if luatex or xelatex
  \ifxetex
    \usepackage{mathspec}
  \else
    \usepackage{fontspec}
  \fi
  \defaultfontfeatures{Ligatures=TeX,Scale=MatchLowercase}
\fi
\IfFileExists{upquote.sty}{\usepackage{upquote}}{}
\IfFileExists{microtype.sty}{%
\usepackage[]{microtype}
\UseMicrotypeSet[protrusion]{basicmath} % disable protrusion for tt fonts
}{}
\PassOptionsToPackage{hyphens}{url} % url is loaded by hyperref
\usepackage[unicode=true]{hyperref}
\hypersetup{
            pdfborder={0 0 0},
            breaklinks=true}
\usepackage{graphicx,grffile}
\makeatletter
\def\maxwidth{\ifdim\Gin@nat@width>\linewidth\linewidth\else\Gin@nat@width\fi}
\def\maxheight{\ifdim\Gin@nat@height>\textheight\textheight\else\Gin@nat@height\fi}
\makeatother
\setkeys{Gin}{width=\maxwidth,height=\maxheight,keepaspectratio}
\IfFileExists{parskip.sty}{%
\usepackage{parskip}
}{% else
\setlength{\parindent}{0pt}
\setlength{\parskip}{6pt plus 2pt minus 1pt}
}
\providecommand{\tightlist}{%
  \setlength{\itemsep}{0pt}\setlength{\parskip}{0pt}}
\ifx\paragraph\undefined\else
\let\oldparagraph\paragraph
\renewcommand{\paragraph}[1]{\oldparagraph{#1}\mbox{}}
\fi
\ifx\subparagraph\undefined\else
\let\oldsubparagraph\subparagraph
\renewcommand{\subparagraph}[1]{\oldsubparagraph{#1}\mbox{}}
\fi

\makeatletter
\def\fps@figure{htbp}
\makeatother

\begin{document}

\title{Discovery of the Hidden State in Ionic Models Using a Domain-Specific
Recurrent Neural Network}

\author{Shahriar Iravanian \thanks{Division of Cardiology,Section of Electrophysiology, Emory University Hospital, Atlanta, GA, USA (shahriar.iravanian@emoryhealthcare.org)}}

%\affil[*]{Division of Cardiology,Section of Electrophysiology, Emory University Hospital, Atlanta, GA, USA}

\date{}

\maketitle

\begin{abstract}

  Ionic models, the set of ordinary differential equations (ODEs) describing the
  time evolution of the state of excitable cells, are the cornerstone of modeling
  in neuro- and cardiac electrophysiology. Modern ionic models can have tens of state
  variables and hundreds of tunable parameters. Fitting ionic models to experimental
  data, which usually covers only a limited subset of state variables, remains a challenging
  problem. In this paper, we describe a recurrent neural network architecture designed
  specifically to encode ionic models. The core of the model is a Gating Neural Network
  (GNN) layer, capturing the dynamics of classic (Hodgkin-Huxley) gating variables.
  The network is trained in two steps: first, it learns the theoretical model coded
  in a set of ODEs, and second, it is retrained on experimental data. The retrained
  network is interpretable, such that its results can be incorporated back into the
  model ODEs. We tested the GNN networks using simulated ventricular action potential
  signals and showed that it could deduce physiologically-feasible alterations of
  ionic currents. Such domain-specific neural networks can be employed in the exploratory
  phase of data assimilation before further fine-tuning using standard optimization
  techniques.

\end{abstract}

\subsection{1. Introduction}\label{introduction}

Ionic models, the set of ordinary differential equations (ODEs)
describing the time evolution of the state of excitable cells, are the
mainstay of cellular and tissue modeling in neuro- and cardiac
electrophysiology. The first ionic model was the two-variable
Hodgkin-Huxley model of the squid giant axon \cite{HODGKIN1952}.
This model was successful in explaining the generation and propagation
of action potentials. Since then, thousands of models have been proposed
for various types of excitable cells in different species \cite{Clayton2011,Fenton:2008}.

The majority of ionic models have retained the general structure of the
Hodgkin-Huxley model while adding a large degree of complexity. For example,
the O'Hara-Rudy human ventricular model has 41 state variables in its most basic
form \cite{Ohara2011}. The large number of parameters raises major concerns
regarding over-fitting, validity, and lack of predictive power. The general
process to develop a new model is to separate and characterize different ionic
currents individually using patch clamping techniques in isolated cells and
then combine all these separate pieces into a coherent model based on the
behavior of the whole cell. One major obstacle to the latter step is that only
one or two of the state variables are actually detectable in tissue experiments,
while other variables remain \emph{hidden}. The primary \emph{observable} state
variable in cardiac and neuronal ionic models is the transmembrane
potential, which can be measured using either micro-electrodes or
optical mapping techniques after staining the tissue with voltage-sensitive
fluorescent dyes \cite{Herron2012}. The second observable in some experimental
setup is the  concentration of intracellular calcium that can be measured using
optical mapping and calcium-sensitive dyes.

Our main problem can be cast as how to deduce and optimize the dynamics
of excitable cells when most of the state variables are hidden. Of
course, this is not a unique situation and occurs repeatedly in system
identification field. Over the years, many different optimization and
probabilistic methods, under the general moniker of data assimilation, have
been adopted to solve hidden-state identification in ionic models
\cite{Smirnov2020,Pouranbarani2019,Krogh-Madsen2016,Cairns2017,Hoffman2016,Barone2017}.

In the last few years, scientific machine learning has emerged as a
powerful alternative to the traditional methodology in the study of
dynamical systems \cite{Roscher2020}. It tries to combine machine learning
methods, such as deep neural networks, with prior domain knowledge encoded in
ODEs and partial differential equations (PDEs). Scientific machine learning is
especially useful in situations when we have an imperfect or poorly
defined physical model but a large amount of experimental data, as is
the case of system identification for ionic models.

According to the universal approximation theorem, it is not surprising
that neural networks can be used to describe dynamical systems.
However, the black-box nature of neural networks is a significant drawback, as
we need an \emph{interpretable} and \emph{explainable} representation of the
model. The keys to a successful application of scientific machine learning are,
first, to be able to encode our prior knowledge (generally in the form of a
system of ODEs) into the model, and second, to be able to interpret the outputs
of the model.

In this paper, our goal is to develop a machine learning architecture
(a domain-specific recurrent neural network) to solve the hidden-variable
identification in ionic models and to help with optimizing model parameters.
Specifically, we like to find the optimized version of an ionic model for a given
time-series of observables. The main motivation for this study was the
challenges we faced trying to apply established ionic models to
experimental data. For example, in one case, we had a large amount of
optical mapping data from guinea pig hearts before and after the infusion of
QT prolonging medications. Our initial plan was to verify the
experimental data by using a standard guinea pig ventricular model \cite{Luo1994}.
However, we noted that baseline simulated data did not match perfectly to the
experimental baseline data. Therefore, we had to modify the model to fit
the observed baseline data before searching the large parameter space to find
a solution to the effects of the medication. In summary, we were faced with
a non-perfect and inexact ionic model, which, at the same time, could not be
ignored, as it encoded the bulk of our prior knowledge of the system of interest.

\subsection{2. Methods}\label{methods}

\subsubsection{2.1 Overview}\label{overview}

We start by describing the motivating example (see above) in a more
formal language. Let $\mathcal{M}$ be an ionic model and let
$\mathbf{X}$ and $\mathbf{Y}$ be two datasets composed of the
observables of $\mathcal{M}$ recorded before and after some
interventions (e.g., application of a membrane-active drug). Our goal is
to find $\mathcal{M}'$, a modification of $\mathcal{M}$, that
explains the effects of the intervention. In theory, $\mathcal{M}$
should reproduce $\mathbf{X}$. In practice, $\mathcal{M}$ is
imperfect, and its output deviates from $\mathbf{X}$. Our solution is
to

\begin{enumerate}
\def\labelenumi{\arabic{enumi}.}
\tightlist
\item
  Train a neural network $\mathcal{N}$ to learn $\mathcal{M}$.
\item
  Retrain $\mathcal{N}$ on the control dataset $\mathbf{X}$ to
  obtain $\mathcal{N}_X$.
\item
  Retrain $\mathcal{N}$ on the treatment dataset $\mathbf{Y}$ to
  obtain $\mathcal{N}_Y$.
\item
  Modify $\mathcal{M}$ based on $\mathcal{N}_X$ and $\mathcal{N}_Y$ to obtain
  $\mathcal{M}_X$ and $\mathcal{M}_Y$ (assuming the neural networks are interpretable).
\item
  Read the effects of the intervention by comparing $\mathcal{M}_X$ and
  $\mathcal{M}_Y$.
\end{enumerate}

In section 2.2, we provide a general overview of the ionic models
($\mathcal{M}$). Sections 2.3 and 2.4 are dedicated to describing
our neural network architecture and the core of its interpretability,
the GNN (Gating Neural Network) layer. Training and retraining methods
are discussed in sections 2.5 and 2.6. Finally, in section 2.7, we show
how to convert neural networks back to neural-ODEs, which are amenable to
integration using standard ODE solvers. The main difficulty lies in the
last step, as neural networks are usually considered to be black-boxes.
As we will explain, our networks are designed with a focus on the
interpretability of their internals.

\subsubsection{2.2 Generic Ionic Models}\label{generic-ionic-models}

In this section, we set up the preliminary definitions and describe the
generic structure of the ionic models before explaining our machine
learning model in the subsequent sections.

Let $\mathbf{u} = \{ u_i \}$ be the state vector, composed of $k$
state variables. For most ionic models, the state variables, $u_i$,
are of three main types: $V_m$, which is the transmembrane potential
and is the main observable, one or more intracellular ion
concentrations, and multiple gating variables.

The system is defined as an ODE,
$\mathbf{u}' = \mathbf{f}(\mathbf{u}, t)$, where prime denotes time
derivation. The dynamics of the gating variables follows the generic
Hodgkin-Huxley formulation. Each gating variable has a value in the
range 0 to 1 and evolves according to (say, for a variable $m$),

\[
  m' = \alpha(V_m) (1 - m) - \beta(V_m) m
  \,,
  \tag{2.2.1}
\]

where $\alpha$ and $\beta$ are the reactivation and deactivation
rates, respectively. The rates are dependent on $V_m$ and, in some
models, on other ionic concentrations, but never on the gating
variables. We will use this fact in the next section to derive a
simplified recurrent network layer. From hereon, we drop the explicit
dependence on $V_m$ to reduce clutter. Eq. 2.2.1 is usually written
as,

\[
  m' = \frac{m_\infty - m}{\tau_m}
  \,,
  \tag{2.2.2}
\]

for

\[
  m_\infty = \frac{\alpha}{\alpha + \beta}
  \,,
  \tag{2.2.3}
\]

and

\[
  \tau_m = \frac{1}{\alpha + \beta}
  \,,
  \tag{2.2.4}
\]

where $m_\infty$ and $\tau$ are the steady-state value and the
time-constant of $m$ and are functions of $V_m$.

The dynamics of $V_m$ is described as

\[
  {V_m}' = \frac{1}{C_m} \left( \sum_j I_j(\mathbf{u}) + I_{\text{stim}}(t)  \right)
  \,,
  \tag{2.2.5}
\]

where $I_j$s are various transmembrane currents, which can depend on
any of the state variables but not directly on $t$, $C_m$ is a
constant quantifying the membrane capacitance, and $I_{\text{stim}}$
is the time-varying external stimulation current.

The fast sodium current (here for the ten Tusscher ventricular model,
see below) is a good example of the form of the equations describing ionic currents,

\[
  I_\text{Na} = g_\text{Na} m ^ 3  h  j \left(V_m - E_\text{Na}(\text{Na}_i, \text{Na}_o) \right)
  \,,
  \tag{2.2.6}
\]

where $m$, $h$, and $j$ are gating variables. In Eq. 2.2.6,
$g_\text{Na}$ is a model parameter and the sodium reversal potential
($E_\text{Na}$) is a function of the intracellular and extracellular
sodium concentrations.

In addition to $V_m$ and the gating variables, $\mathbf{u}$ also
includes various intracellular concentrations. The general ODE for a
typical concentration (say $c$) is

\[
    c' = f_c(c, I_c)
    \,,
    \tag{2.2.7}
\]

where $I_c$ is the current corresponding to $c$. In contrast to
classic gating variables, $I_c$, and hence $c'$, may depend on
gating variables.

In summary, an ionic model is composed of a state variable
$\mathbf{u}$, the initial values, $\mathbf{u}(0)$, and a set of ODEs
similar to Eqs. 2.2.2, 2.2.5, and 2.2.7. In many cardiac applications, a
one-, two- or three-dimensional Laplacian of $V_m$ is added to the
right side of Eq. 2.2.5 to form a system of PDEs. In this paper, we
limit ourselves to the study of the ODE form of the ionic models.

A common feature of the ionic models, especially the cardiac ones, is
the presence of two or more time scales. Usually, $V_m$ and some gating
variables, like $m$ and $h$, are fast; whereas, ionic concentrations
and some other gating variables are slow (tens to hundreds of
milliseconds). Specially, $m_\infty$ and $\tau$ in Eq. 2.2.2 can be
considered constant for the duration of one time step
($\Delta{t} \lesssim 1\,\text{ms}$). Therefore, we can explicitly integrate
Eq. 2.2.2 to obtain

\[
  m(t + \Delta{t}) = m_\infty + \left( m(t) - m_\infty \right) e^{-\Delta{t}/\tau}
  \tag{2.2.8}
  \,.
\]

This trick is called the Rush-Larsen method and forms the core of our recurrent
network (see section 2.4) \cite{Rush1978}.

\subsubsection{2.3 General Network
Architecture}\label{general-network-architecture}

The key ingredient of scientific machine learning is the fusion of deep learning
techniques with traditional scientific knowledge, usually encoded as systems of
ODEs and PDEs. Here, we limit our discussion to initial-value ODEs,

\[
  \mathbf{u}' = \mathbf{f}(\mathbf{u}, t; \theta)
  \tag{Eq. 2.3.1}
  \,,
\]

where $\mathbf{u}$ is the state vector, $t$ is time, $\theta$ represents model
parameters, and the initial condition, $\mathbf{u}(0)$ is known. A general
framework to mix ODEs with machine learning is provided by the concept of
universal ordinary differential equations (UODE) \cite{rackauckas2020}:

\[
  \mathbf{u}' = \mathbf{f}(\mathbf{u}, t, U_{\theta'}(\mathbf{u}, t); \theta)
  \tag{Eq. 2.3.2}
  \,.
\]

The machine learning part is provided by $U_{\theta'}(u, t)$ representing a
universal approximator -- a neural network, a random forest, a Chebychev
expansion, or any other suitable approximator. In a UODE, $U_{\theta'}(u, t)$
replaces part or the entirety of the ODE derivative function. For the rest of
the paper, we assume that $U_{\theta'}$ is a neural network, therefore

\[
  \mathbf{u}' = \mathbf{f}(\mathbf{u}, t, \text{NN}_{\theta'}(\mathbf{u}, t); \theta)
  \tag{Eq. 2.3.3}
  \,.
\]

One of the main challenges of building a UODE model is at the interface of the
neural network, NN, and the ODE when NN require access to the ODE
estimated error during training. One possible solution is to use ODE sensitivity
analysis (using adjoint equations) to provide an estimate of the error that then
back-propagates into NN. Another challenge in applying UODE methodology
to solving ionic models is that most of the variables in the state vector remain
unobservable. In Eqs. 2.3.1-2.3.3, it is implicitly assumed that $\mathbf{u}$ is observable.
As mentioned above, in electrophysiology (and many other fields), only a limited
subset of the state variables is observable, and the rest is hidden. We address
this problem by using recurrent neural networks (RNN). A typical feed-forward
neural network is stateless; however, we can make it stateful by adding recurrent
layers. The resulting RNN is better equipped in representing dynamical maps that
flows. Therefore, we modify our ODE system to a map,

 \[
  \mathbf{v}(t+\Delta{t}) = \mathbf{f}(\mathbf{v}, t, \text{RNN}_{\mathbf{\theta'},\mathbf{h}}(\mathbf{v}, t); \theta)
  \tag{Eq. 2.3.4}
  \,,
\]

where $\mathbf{v} \in \mathbf{u}$ is the set of observables and $\mathbf{h}$
stands for the hidden state. In essence, the neural network becomes an ODE
integrator. We can make another simplification by absorbing the entirety of the
ODE system into the recurrent network by setting $\mathbf{f}$ to the identity
function, and $\theta' = \theta$ to get

\[
  \mathbf{v}(t+\Delta{t}) = \text{RNN}_{\mathbf{\theta},\mathbf{h}}(\mathbf{v}, t)
  \tag{Eq. 2.3.5}
  \,.
\]

Eq. 2.3.5 forms the basis of our neural network. The most commonly used recurrent
layer is the long short-term memory or LSTM layer \cite{Hochreiter1997,Gers2000}.
Although LSTM has been very successful in general applications, it lacks easy
interpretability. Therefore, we simplified and modified the LSTM into the GNN
(described in section 2.4), a recurrent layer that is specifically
optimized for Hodgkin-Huxley style gating variables. However, not all the state
variables in ion models are classic gating variables. We incorporate an LSTM
layer in our network to handle these variables, accepting the trade-off of
losing interpretability for them.

Figure 1 depicts the schematic of our network. The network is composed of two
sub-networks: $\mathcal{N}_1$, which is $\phi 1$ plus a GNN layer, and
$\mathcal{N}_2$, which is formed from $\phi 2$, an LSTM layer, and $\phi 3$.
$\mathcal{N}_1$ handles the classic gating variables, and therefore uses a GNN
layer. $\mathcal{N}_2$ is responsible for other state variables (non-classic
gates, concentrations, and $V_m$) with the help of its LSTM layer. The other
layers $(\phi 1, \phi 2, \phi 3$) are standard dense and fully-connected one or
two-layer deep networks.

\begin{figure}
\centering
\includegraphics{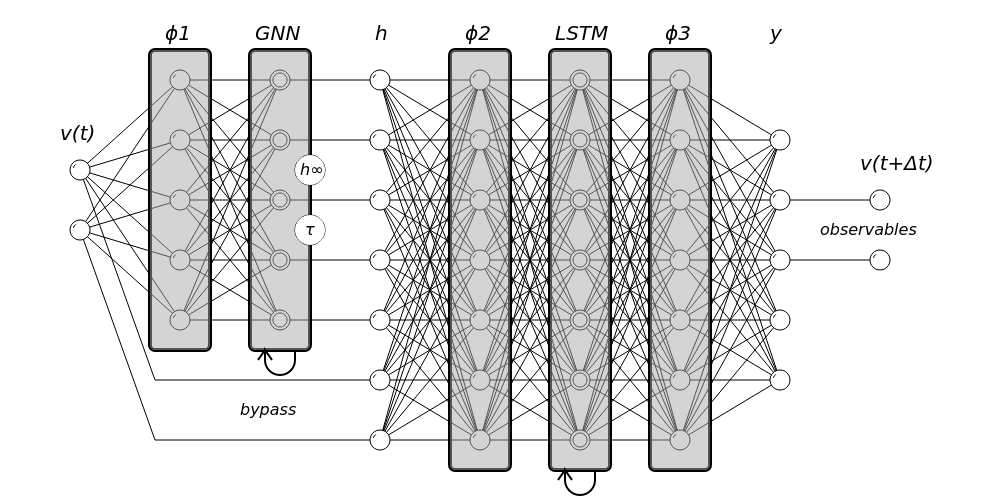}
\caption{The general schematic of our neural network architecture. The network is composed of
two sub-networks: $\mathcal{N}_1$, formed by $\phi 1$ (two fully-connected dense layers) and GNN
(a domain-specific recurrent layer), and $\mathcal{N}_2$, formed by $\phi 2$, $\phi 3$, and
a recurrent layer (LSTM). }
\end{figure}

The entire network transforms the observables at time $t$, i.e., $\mathbf{v}(t)$,
to the observables at $t + \Delta{t}$. However, the output of $\mathcal{N}_1$,
which corresponds to the gating variables, is exposed and is used for training.
Note that the input, $\mathbf{v}(t)$, is fed to both $\mathcal{N}_1$ and, via
a bypass, to $\mathcal{N}_2$.

\subsubsection{2.4 Gating Neural Network
(GNN)}\label{gating-neural-network-gnn}

The GNN is a recurrent layer designed to \emph{integrate} Hodgkin-Huxley gating
state variables based on Eq. 2.2.8. The goal is the ease of training and
interpretability. Each GNN layer stores a state vector, $\mathbf{h}(t)$, which
generally corresponds to the gating variables of $\mathbf{u}$.

At each time step, $\mathbf{h}(t)$ is updated. We need the steady-state value,
$\mathbf{h}_\infty(t)$, and the time constant, $\mathbf{\tau}_h(t)$, to solve
the right hand side of Eq. 2.2.8. For classic gating variables, both
$\mathbf{h}_\infty(t)$ and $\mathbf{\tau}_h(t)$ depend on $V_m$, and rarely, ionic
concentrations, but never on other gating variables. Therefore, we calculate
$\mathbf{h}_\infty(t)$ as

\[
  \mathbf{h}_\infty(t) = \sigma (\mathbf{W}_\infty \mathbf{x}(t) + \mathbf{b}_\infty)
  \tag{Eq. 2.4.1}
  \,,
\]

where $\mathbf{W}_\infty$ stands for the layer weights, $\mathbf{b}_\infty$ is bias,
and $\mathbf{x}(t)$ is the input to the GNN layer (the output of $\phi 1$
in Figure 1). We have chosen a sigmoid transfer function, $\sigma$, to enforce
the 0 to 1 range for the gating variables and their steady-state values. Note
the lack of a self-loop, i.e., $\mathbf{h}_\infty$ does not directly depend on
$\mathbf{h}$. This is in contrasts to the LSTM, where the equation counterpart to
2.4.1 includes an additional term $\mathbf{W} . \mathbf{h}(t)$.

In contrast to $\mathbf{h}_\infty$, $\mathbf{\tau}_h$ is not necessarily in
the range $[0,1]$ (in fact, it has unit of time). Recognizing that $0 < e^{-\Delta{t}/\tau} < 1$
for $\tau > 0$, we define the update rate as

\[
  \mathbf{\rho}_h(t) = e^{-\Delta{t}/\mathbf{\tau}_h(t)}
  \tag{Eq. 2.4.2}
  \,.
\]

Similar to Eq. 2.4.1,

\[
  \mathbf{\rho}_h(t) = \sigma (\mathbf{W}_\tau \mathbf{x}(t) + \mathbf{b}_\tau)
  \tag{Eq. 2.4.3}
  \,,
\]

for weight matrix and bias $\mathbf{W}_\tau$ and $\mathbf{b}_\tau$. Combining
Eqs. 2.4.1 and 2.4.3 with Eq. 2.2.8, we obtain the GNN update formula,

\[
  \mathbf{h}(t + \Delta{t}) = \mathbf{\rho}_h(t) \mathbf{h}(t) + \left(\mathbf{1} - \mathbf{\rho}_h(t) \right) \mathbf{h}_\infty(t)
  \tag{Eq. 2.4.4}
  \,.
\]

\subsubsection{2.5 First-Pass Training}\label{first-pass-training}

During the first-pass training, $\mathcal{N}$ has full access to the
\emph{hidden} states and learns $\mathcal{M}$.

To prepare the training dataset, we start with Eq. 2.3.1 (repeated here),

\[
  \mathbf{u}' = \mathbf{f}(\mathbf{u}, t; \theta)
  \tag{Eq. 2.5.1}
  \,.
\]

$\mathcal{M}$ includes $\mathbf{f}$, $\mathbf{\theta}$, and the initial condition,
$\mathbf{u}(0)$. Let $\mathbf{u}_{i}$ be the solution to 2.5.1 at $t = i \Delta{t}$.
In practice, we solve the model for a range of cycle lengths. Depending on the context,
$\mathbf{u}_{i}$ is either a $k$-element state vector or a $k \times l$ matrix of
$l$ column state vectors, where $l$ is the number of the cycle lengths used for
training.

We introduce three helper functions to select a subset of the state variables from
$\mathbf{u}$: $O(\mathbf{u})$ extracts the observables, $H(\mathbf{u})$ lists the
hidden gating variables handled by the GNN layer, and $\tilde H(\mathbf{u})$ lists
the remaining variables processed by the LSTM layer. By definition, $H(\mathbf{u}) \cap \tilde H(\mathbf{u}) = \emptyset$ and $H(\mathbf{u}) \cup \tilde H(\mathbf{u}) = \mathbf{u}$.

The data should be presented in a sequential manner to a recurrent network. Hence,
we train our model by feeding the data in $(\mathbf{u}_0,\mathbf{u}_1),(\mathbf{u}_1, \mathbf{u}_2),\cdots$ order, where the
first item of each ordered list is the input to $\mathcal{N}$ and the second
item is the expected output.

The observables generally do not contain enough information for the trained network
to capture the model faithfully. We need to explicitly enforce the constraints encoded
in $\mathcal{M}$ by including the value of the hidden variables in the loss function.

Let $\mathcal{N}$ be the entire network in Figure 1. It is composed of two sub-networks,
$\mathcal{N}_1$ and $\mathcal{N}_2$. We train $\mathcal{N}_1$ by feeding the observables,
$O(\mathbf{u}_i)$, as input, and compare its output to the value of the gating variables
in the next timestep. This adds a term proportional to $L_2(\mathcal{N}_1(O(\mathbf{u}_i)), H(\mathbf{u}_{i+1}))$ to the loss function, where $L_2$ stands for the square-error
loss function.

Similarly, $\mathcal{N}_2$ is trained by adding a term $L_2(\mathcal{N}_2(O(\mathbf{u}_i)), \tilde H(\mathbf{u}_{i+1}))$ to the loss function. Note that $\mathcal{N}_2$ implicitely
takes the output of $\mathcal{N}_1$ as part of its input.

The final loss function is the combination of the $L_2$ terms and a standard regularization
term,

\[
  L(\mathbf{u}_i, \mathbf{u}_{i+1}) =
    L_2(\mathcal{N}_1(O(\mathbf{u}_i)), H(\mathbf{u}_{i+1})) +
    L_2(\mathcal{N}_2(O(\mathbf{u}_i)), \tilde H(\mathbf{u}_{i+1})) +
    \lambda L_2(W)
  \tag{Eq. 2.5.2}
  \,,
\]

where $W$ stands for all the wights in $\mathcal{N}$, and $\lambda$ is a super-parameter,
controlling the influence of the regularization term.

\subsubsection{2.6 Second-Pass Retraining (Transfer
Learning)}\label{second-pass-retraining-transfer-learning}

The network learns the theoretical model, $\mathcal{M}$, in the first-pass. For
the second-pass, the goal is to fine-tune the model based on experimental
recordings. Let $\mathbf{v}_i = O(\mathbf{u}_i)$
be the observables at $t = i \Delta{t}$. The training dataset is
$\mathbf{X} = (\mathbf{v}_{0}, \mathbf{v}_{1}, \mathbf{v}_{2},\cdots)$.

The first-pass/second-pass training architecture is a form of \emph{transfer learning}.
We pre-train the network on an easier or more accessible dataset first and then
modify it to learn the target dataset. Here, we keep $\mathcal{N}_2$ and $\phi 1$
frozen and only retrain the GNN layer. Let $\mathcal{N}_{1,X}$ be $\mathcal{N}_1$
retrained on $\mathbf{X}$. Similarly, $\mathcal{N}_X$ is the new network made of
$\mathcal{N}_{1,X}$ and $\mathcal{N}_2$, and $W_X$ and $L_X$ are the new versions
of $L$ and $W$. Since  $\mathcal{N}_2$ is frozen, we have $\mathcal{N}_{2,X} = \mathcal{N}_{2}$.

We start with a standard loss function based on the input and the expected output
of $\mathcal{N}_X$:

\[
  L_X(\mathbf{v}_i, \mathbf{v}_{i+1}) = L_2\left(\mathcal{N}_X(\mathbf{v}_i), \mathbf{v}_{i+1} \right) + \lambda L_2(W_X)
  \tag{Eq. 2.6.1}
  \,.
\]

The problem with Eq. 2.6.1 is that $\mathcal{N}_X$ can drift widely from $\mathcal{N}$,
such that the resulting network may become uninterpretable. The solution is to add
a term to the loss function to penalize the drift by linking the outputs of $\mathcal{N}_1$
and $\mathcal{N}_{X,1}$,

\[
  L_X(\mathbf{v}_i, \mathbf{v}_{i+1}) =
    L_2\left(\mathcal{N}_X(\mathbf{v}_i), \mathbf{v}_{i+1} \right) +
    \eta L_1\left(\mathcal{N}_{1,X}(\mathbf{v}_i), \mathcal{N}_1(\mathbf{v}_i) \right) +
    \lambda L_2(W_X)
  \tag{Eq. 2.6.2}
  \,,
\]

where we introduce an additional super-parameter, $\eta$, which controls how much
the network drifts during retraining. Finding the right value of $\eta$ is critical
in getting correct and interpretable results. The $L_1$ norm in the $\eta$-term
promotes a sparse solution, such that most gates stay the same, and only one or
few of them are changed.

\subsubsection{2.7 Neural ODE}\label{neural-ode}

It is possible to convert a trained or retrained GNN network into a neural
ODE and solve it using standard ODE solvers. As mentioned above, the GNN network
is designed as an integrator and can theoretically solve the model. However,
it is akin to an explicit ODE solver (the GNN layer is based on the Rush-Larsen method,
an explicit exponential ODE solver). Most ionic models are numerically stiff and
need an implicit solver for stable integration. By converting the network into a formal
neural-ODE, one can apply optimized implicit solvers to the resulting model.

The resulting ODE follows Eq. 2.3.4, where the neural network is stateless, and the
state vector ($\mathbf{u}$) resides outside the network. Therefore, to convert a stateful
recurrent network into a stateless neural network, we need to strip away any part
of the network that depends on its internal state vectors. What remains is $\phi 1$
and the trained weights in the GNN layer.

For each input $\mathbf{u}$, we feed $\mathbf{v} = O(\mathbf{u})$ into $\phi 1$ and
then calculate $\mathbf{h}_\infty$ and $\mathbf{\rho}_h$ using Eqs. 2.4.1 and 2.4.3.
Next, we calculate $\mathbf{\tau}_h$ by inverting Eq. 2.4.2,

\[
  \mathbf{\tau}_h = -\frac{\Delta{t}}{\log{\mathbf{\rho}_h}}
  \tag{Eq. 2.7.1}
  \,.
\]

According to 2.2.2,

\[
  \mathbf{h}' = \frac{\mathbf{h}_\infty - \mathbf{h}}{\mathbf{\tau}_h}
  \tag{Eq. 2.7.2}
  \,,
\]

where $\mathbf{h} = H(\mathbf{u})$. We can then combine $\mathbf{h}'$ with the rest
of the model ODEs to find a combined ODE/neural network, as in Eq. 2.3.4.

\subsection{3 Results}\label{results}

We use the \emph{ten Tusscher, Noble, Noble, Panfilov} human ventricular ionic
model as our example (ten Tusscher from here on) \cite{tenTusscher2004}. The examples
are coded in the Julia programming language \cite{Bezanson2017} with the help of \textbf{DifferentialEquations.jl} ODE solving library \cite{rackauckas2017}
and \textbf{Flux.jl} neural network framework \cite{Flux.jl-2018}. The GNN model
and the examples are available from https://github.com/shahriariravanian/gnn.

The ten Tusscher model is a moderate size ionic model with good numerical stability.
This model has seventeen state variables: $V_m$, four ion concentrations, and twelve
gates. Ten gates are classic Hodgkin-Huxley gates. These gates and the corresponding
currents are:

\begin{enumerate}
\def\labelenumi{\arabic{enumi}.}
\tightlist
\item
  Fast sodium current ($I_\text{Na}$): gates $m$, $h$, and $j$.
\item
  L-type calcium current ($I_\text{Ca,L}$): classic gates $d$ and
  $f$, and atypical gate $f_\text{Ca}$.
\item
  Rapid delayed rectifier potassium current ($I_\text{Kr}$):
  $x_{r1}$ and $x_{r2}$.
\item
  Slow delayed rectifier potassium current ($I_\text{Ks}$): $x_{s}$.
\item
  Inward rectifier potassium current ($I_\text{K1}$): non-state gate
  $K_1$.
\item
  Transient outward potassium current ($I_\text{to}$): gates $r$ and
  $s$.
\item
  The sarcoplasmic reticulum calcium-induced calcium-release current
  ($I_\text{rel}$): atypical gate $g$.
\end{enumerate}

The following gates are encoded in the GNN layer: $m$, $h$, $j$, $d$, $f$,
$x_{r1}$, $x_{r2}$, $x_s$, $r$, and $s$.

\begin{figure}
\centering
\includegraphics{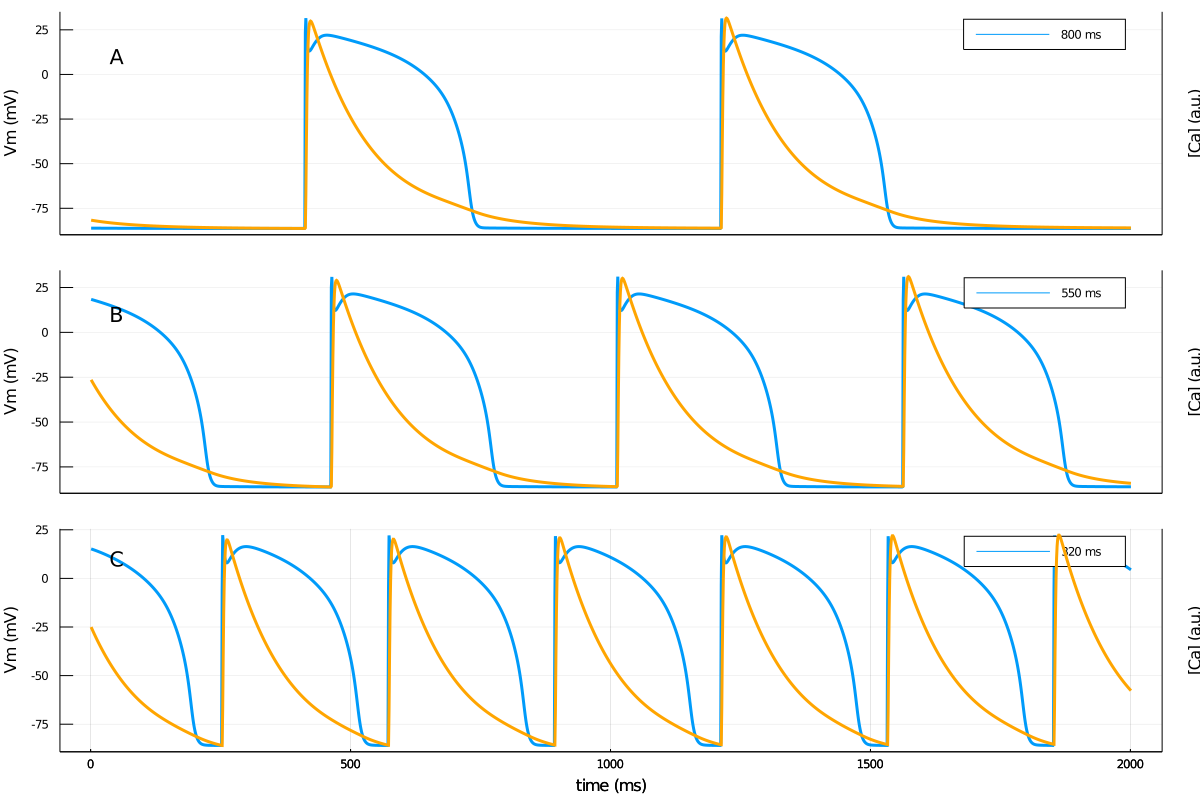}
\caption{The observables for the ten Tusscher model at different pacing cycle lengths.
The blue curve is the transmembrane potential ($V_m$), and the orange curve shows the intracellular
calcium concentration ($Ca_i$).
}
\label{signal_base}
\end{figure}

We integrated the ten Tusscher model for 20 seconds at cycle lengths 300 to 800 ms
in 5 ms increments to generate 101 training data segments. The first half of each
20-second segment was discarded due to transient behavior. The output signal was re-sampled
at 1 ms timestep. Of the 101 segments, 76 were randomly selected as the training
dataset, and the other 25 were used for validation. Figure \ref{signal_base} shows
three examples of the training signals at cycle lengths 320, 550, and 800 ms.

\begin{figure}
\centering
\includegraphics{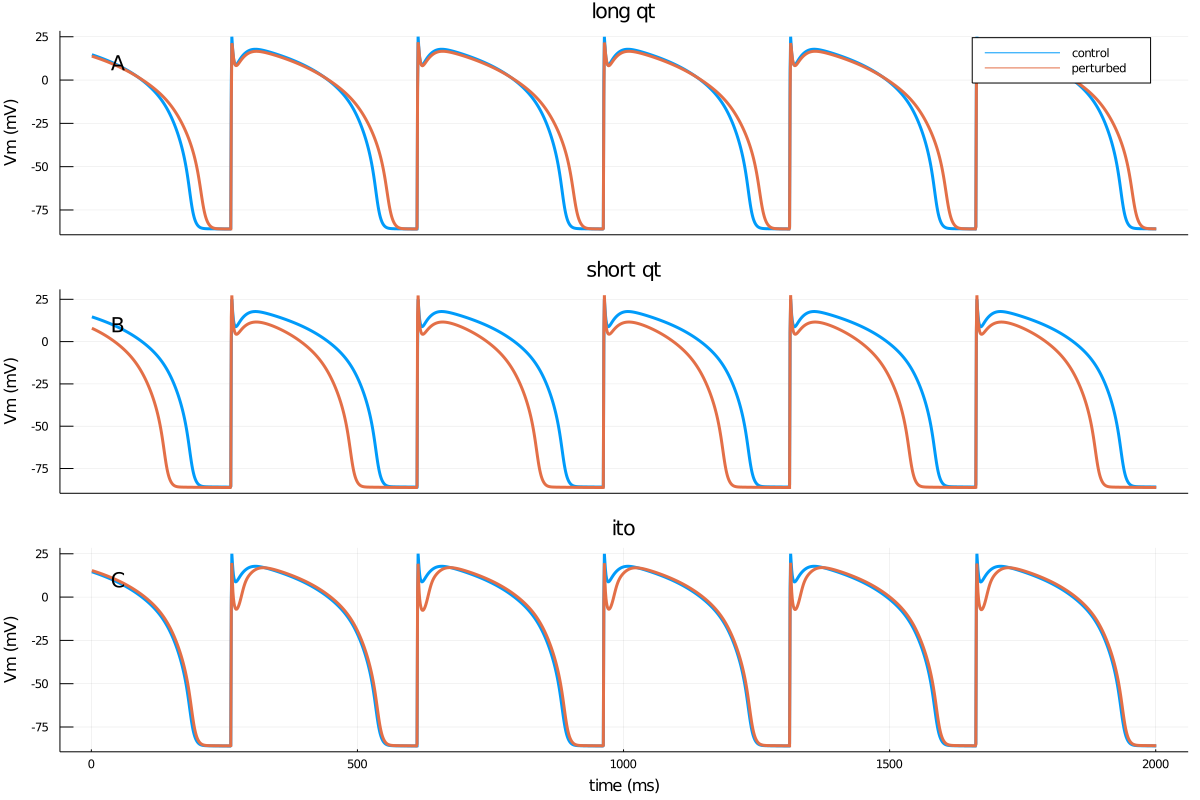}
\caption{Transmembrane potential in different scenarios used to test the model.
The blue curves are the baseline $V_m$. The red curves show the perturbed
signals. }
\label{signal_perturbed}
\end{figure}

In addition, to test the utility of the method to identify the underlying dynamics,
we generated three \emph{perturbed} datasets, corresponding to prolonged APD (\textbf{long qt}), shortened APD (\textbf{short qt}), and increased transient outward current (\textbf{ito})
conditions (Figure \ref{signal_perturbed}).

First-pass training of the neural network was done on the base dataset, where we
assumed the network had access to the full state vector, including all the hidden
variables. All the layers in $\phi 1$, $\phi 2$, and $\phi 3$ had
size 30. As mentioned above, the GNN layer encoded the ten classic gates. The LSTM
layer was tasked with handling the remaining seven variables. A value of
$\lambda = 10^{-4}$ was used for regularization.

Second-pass training was performed only on the observables: $V_m$ and $\text{Ca}_i$.
Considering that we didn't know the optimal value of $\eta$ a priori, a range of values
from $10^{-4}$ to $2\times10^{-3}$ were tried.

\begin{figure}
\centering
\includegraphics{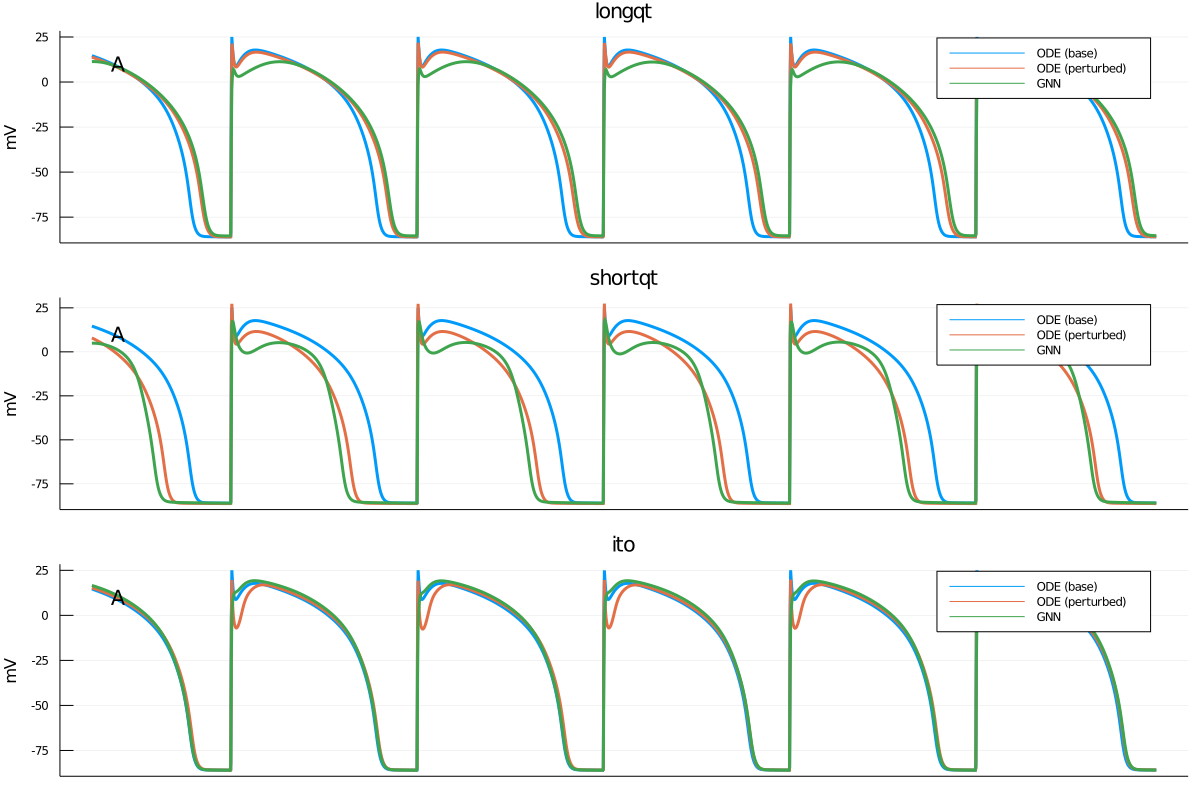}
\caption{ Comparison between the output of the neural ODE, based on the retrained
GNN networks, and the output of the standard ODEs representing the models. The blue
and red curves are the same as in Figure \ref{signal_perturbed}. The green curve is
the output of the neural ODE. }
\label{neural_ode}
\end{figure}

Figure \ref{neural_ode} depicts the base signal (blue, calculated using the standard ODE),
the perturbed signal (red), and the output of the retrained network (green, using a neural ODE
as in 2.7). The results suggest that the GNN network can capture the
essence of the changes in the signals (e.g., APD prolongation in the long qt scenario).
However, the fit between the actual signal and the output of the retrained networks is not
perfect. In what follows, we discuss how the network adjusted the GNN layer weights
for different scenarios.

\subsubsection{3.1 Long QT}\label{long-qt}

We simulated the long QT syndrome (prolonged repolarization) by reducing $I_\text{Kr}$
peak conductance by 50\%, while leaving the gating dynamics constant. This is a
common mechanism for various medications to cause acquired long QT syndrome, which
is pro-arrhythmic, potentially fatal, and of significant clinical importance.

\begin{figure}
\centering
\includegraphics{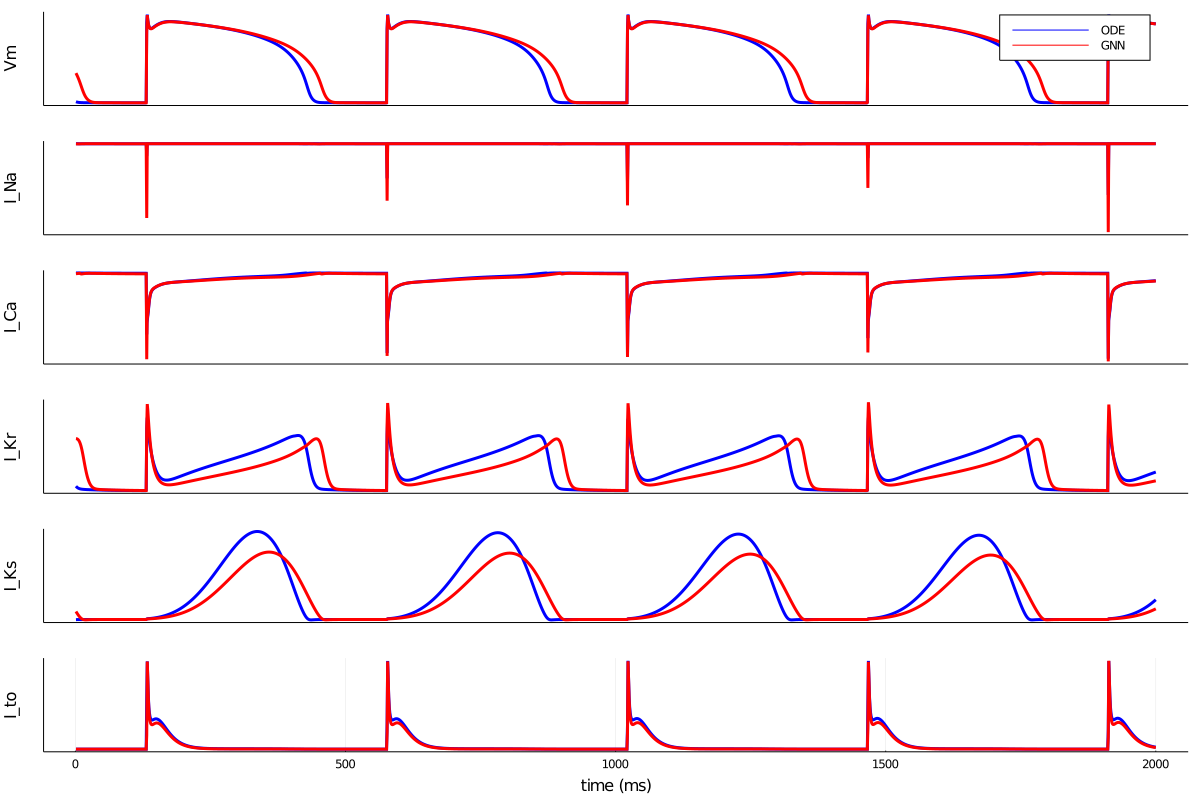}
\caption{ Comparison of the ionic currents and transmembrane potential predicted
by the GNN networks retrained on the control (blue curves) and the perturbed (long
qt scenario) observables (red curves). Note the reduction in both $I_\text{Kr}$
and $I_\text{Ks}$ currents. }
\label{currents_longqt}
\end{figure}

The resulting signal is shown in Figure \ref{signal_perturbed}A. Figure \ref{currents_longqt}
depicts ionic currents after retraining the network with the new dataset.
The currents are calculated by plugging the values of the gating variables obtained
from the retrained network into the corresponding equations of the ten Tusscher model.
For example, to calculate the fast sodium current, we plugged $m$, $h$, $j$ from
$\mathcal{N}_1$ into Eq. 2.2.6.

The GNN network correctly deduced that reduction in potassium currents prolonged
the repolarization phase. In Figure \ref{currents_longqt}, both $I_\text{Kr}$ and
$I_\text{Ks}$ are decreased; whereas, the signal was generated by decreasing only
$I_\text{Kr}$. The ten Tusscher model, with 17 variables, has a large
amount of redundancy. The observables alone do not contain enough information
to completely constraint the model. Therefore, an exact match between the underlying
mechanism of signal generation and what the network predicts is not attainable.

\subsubsection{3.2 Short QT}\label{short-qt}

\begin{figure}
\centering
\includegraphics{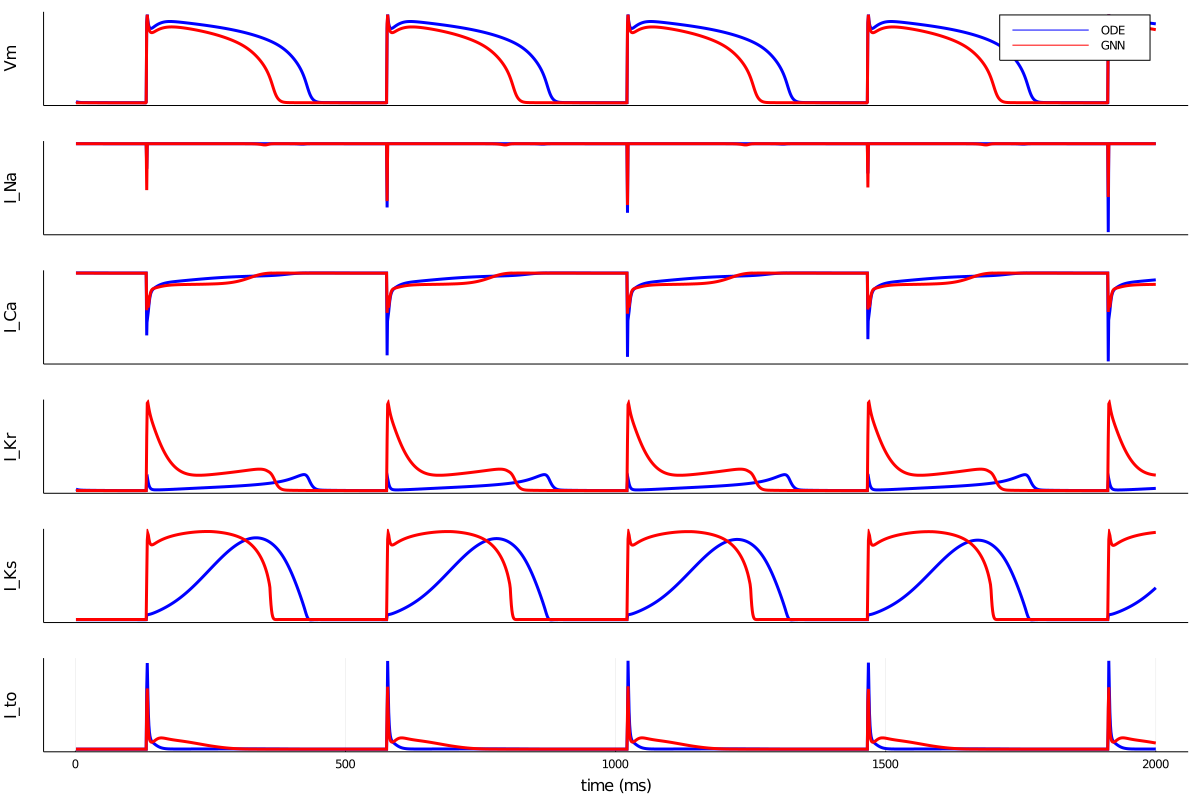}
\caption{ Comparison of the ionic currents and transmembrane potential predicted
by the GNN networks retrained on the control (blue curves) and the
perturbed (short qt scenario) observables (red curves). Note that both
$I_\text{Kr}$ and $I_\text{Ks}$ currents are increased and shifted to the left. }
\label{currents_shortqt}
\end{figure}

The short qt scenario was generated by decreasing $I_\text{Ca,L}$ by 50\% (Figure \ref{signal_perturbed}B). The retrained model currents are shown in Figure
\ref{currents_shortqt}. The model achieved a short APD by increasing the
potassium currents, which is a mechanistically valid solution. In general, the APD
is determined by the balance between calcium and potassium currents during phase 2
of action potentials. Therefore, a short APD is a sign of either reduced calcium
currents or increased potassium currents. Again, the observables lack enough information
to break the symmetry between the two options.

\subsubsection{3.3 Increased Ito}\label{increased-ito}

For the third scenario, we increased the $I_\text{to}$ current three-fold. As a
result, phase 1 of the action potential (the notch after the upstroke) became
deeper. This was a difficult test for the model for two reasons. First, phase 1
is very short in duration compared to other phases of the action potential. Second,
in contrast to the other two scenarios, we increased (rather than decreased) a
channel conductance here. The presence of $\sigma$ activation functions in Eqs. 2.4.1
and 2.4.3 prevents the model from increasing the values of any gate above 1.
Moreover, the actual conductance is not part of retraining (it is implicitly encoded
in $\phi 2$). One option would have been to augment the neural network by adding
an affine layer after the GNN layer. However, we decided to keep the model simple
and observe how it could handle the third scenario within the constraints.

\begin{figure}
\centering
\includegraphics{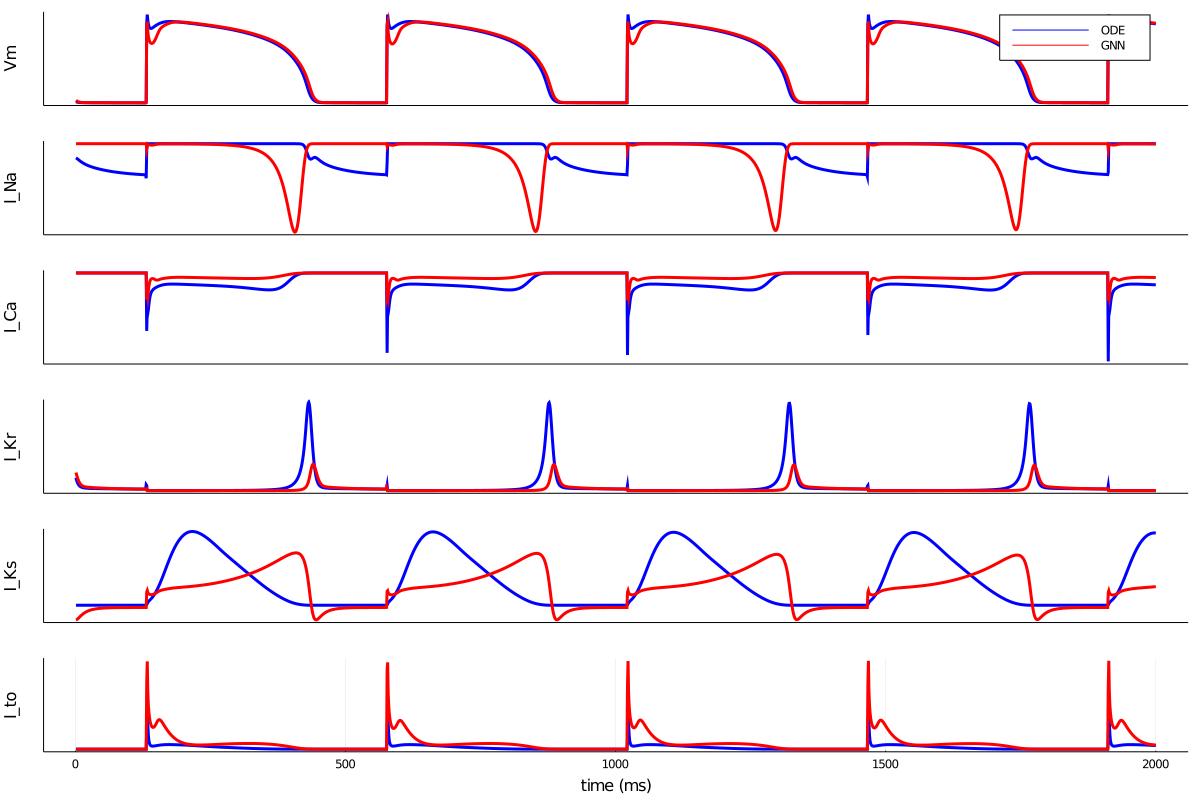}
\caption{ Comparison of the ionic currents and transmembrane potential predicted
by the GNN networks retrained on the control (blue curves) and the perturbed (ito
scenario) observables (red curves). Note the significant increase in the $I_\text{to}$
current. Refer to the text for the discussion of other currents. }
\label{currents_ito}
\end{figure}

The resulting currents are shown in Figure \ref{currents_ito}. The network correctly
identified $I_\text{to}$ as the main contributor and adjusted it accordingly.
Other potassium currents were also modified to compensate for the increased current
through $I_\text{to}$. Interestingly, $I_\text{CaL}$ was reduced during phase 2 of
action potentials to tip the balance toward potassium currents. However, we also
observed undesirable changes in the calcium and, especially, sodium currents.

\subsection{4 Discussion}\label{discussion}

In this paper, we demonstrated the feasibility of optimizing complex cardiac and
neuronal ionic models trained on limited data sets (containing only observable
subsets of state variables) with the help of domain-specific recurrent neural
networks. We designed the network to capture the dynamics of the system. This
allowed us to train the neural network to learn the theoretical model, represented
by a system of ODEs, and interpret the results after the network was retrained
on observables.

Historically, fine-tuning ionic models has been done either in an ad hoc fashion
by manual trial and error or in a more systemic way by utilizing various optimization
techniques, such as ODE sensitivity analysis and population methods \cite{Smirnov2020,Pouranbarani2019,Krogh-Madsen2016,Cairns2017,Hoffman2016,Barone2017,BrittonE2098}.
We believe that our methods, or more generally domain-specific machine-learning
methodology, augments but does not replace the traditional methodology. In traditional
optimization, one selects a limited group of parameters, e.g., the peak ionic conductance,
to optimize against the experimental data, whereas, when using neural networks,
the parameter space is much larger and includes the general shapes of the functions
describing the details of the model. Therefore, it is reasonable to use the domain-specific
neural networks in the exploratory phase to determine the general shape of the functions
and then apply standard optimization techniques for further fine-tuning.

We presented a motivating example in the Introduction about the sub-optimal match
between an established ionic model and experimental data. As discussed in the Results
section, we explored the utility of the network with a GNN layer in solving
the mismatch problem. The GNN-based network was able to learn the model and morph
to the observable data. We were able to interpret the network after it was retrained
on the perturbed data series. The interpretations were physiologically valid but different
from the way the signals were generated in the first place. The main reason for
this difference is that ionic models have many parameters and are generally
over-determined based on experimentally available data. Linking the control and
treated networks by the $\eta$-term in Eq. 2.6.2 restricted the parameter-space.
However, even the restricted space was large enough to prevent a unique solution.
One way to further limit the parameter-space is to impose additional constraints
on the model. For example, if we know that a given drug X is an $I_\text{Kr}$
blocker, it makes sense to add a weighting-factor to the $\eta$-term to encourage
the model to preferentially modify $x_{r1}$ and $x_{r2}$ gates.

\bibliographystyle{plain}
\bibliography{gnn}

\begin{thebibliography}{10}

\bibitem{Barone2017}
Alessandro Barone, Flavio Fenton, and Alessandro Veneziani.
\newblock Numerical sensitivity analysis of a variational data assimilation
  procedure for cardiac conductivities.
\newblock {\em Chaos}, 27:093930, 9 2017.

\bibitem{Bezanson2017}
Jeff Bezanson, Alan Edelman, Stefan Karpinski, and Viral~B. Shah.
\newblock Julia: A fresh approach to numerical computing.
\newblock {\em SIAM Review}, 59:65--98, 1 2017.

\bibitem{BrittonE2098}
Oliver~J. Britton, Alfonso Bueno-Orovio, Karel Van~Ammel, Hua~Rong Lu, Rob
  Towart, David~J. Gallacher, and Blanca Rodriguez.
\newblock Experimentally calibrated population of models predicts and explains
  intersubject variability in cardiac cellular electrophysiology.
\newblock {\em Proceedings of the National Academy of Sciences},
  110(23):E2098--E2105, 2013.

\bibitem{Cairns2017}
Darby~I. Cairns, Flavio~H. Fenton, and E.~M. Cherry.
\newblock Efficient parameterization of cardiac action potential models using a
  genetic algorithm.
\newblock {\em Chaos}, 27:093922, 9 2017.

\bibitem{Clayton2011}
R.~H. Clayton, O.~Bernus, E.~M. Cherry, H.~Dierckx, F.~H. Fenton, L.~Mirabella,
  A.~V. Panfilov, F.~B. Sachse, G.~Seemann, and H.~Zhang.
\newblock Models of cardiac tissue electrophysiology: Progress, challenges and
  open questions, 1 2011.

\bibitem{Fenton:2008}
F.~H Fenton and E.~M. Cherry.
\newblock {M}odels of cardiac cell.
\newblock {\em Scholarpedia}, 3(8):1868, 2008.

\bibitem{Gers2000}
Felix~A. Gers, Jürgen Schmidhuber, and Fred Cummins.
\newblock Learning to forget: Continual prediction with lstm.
\newblock {\em Neural Computation}, 12:2451--2471, 10 2000.

\bibitem{Herron2012}
Todd~J. Herron, Peter Lee, and José Jalife.
\newblock Optical imaging of voltage and calcium in cardiac cells \& tissues, 2
  2012.

\bibitem{Hochreiter1997}
Sepp Hochreiter and Jürgen Schmidhuber.
\newblock Long short-term memory.
\newblock {\em Neural Computation}, 9:1735--1780, 11 1997.

\bibitem{HODGKIN1952}
A~L HODGKIN and A~F HUXLEY.
\newblock A quantitative description of membrane current and its application to
  conduction and excitation in nerve.
\newblock {\em The Journal of physiology}, 117:500--544, 8 1952.

\bibitem{Hoffman2016}
M.~J. Hoffman, N.~S. LaVigne, S.~T. Scorse, F.~H. Fenton, and E.~M. Cherry.
\newblock Reconstructing three-dimensional reentrant cardiac electrical wave
  dynamics using data assimilation.
\newblock {\em Chaos}, 26, 1 2016.

\bibitem{Flux.jl-2018}
Michael Innes, Elliot Saba, Keno Fischer, Dhairya Gandhi, Marco~Concetto
  Rudilosso, Neethu~Mariya Joy, Tejan Karmali, Avik Pal, and Viral Shah.
\newblock Fashionable modelling with flux.
\newblock {\em CoRR}, abs/1811.01457, 2018.

\bibitem{Krogh-Madsen2016}
Trine Krogh-Madsen, Eric~A. Sobie, and David~J. Christini.
\newblock Improving cardiomyocyte model fidelity and utility via dynamic
  electrophysiology protocols and optimization algorithms.
\newblock {\em Journal of Physiology}, 594:2525--2536, 5 2016.

\bibitem{Luo1994}
C~H Luo and Y~Rudy.
\newblock A dynamic model of the cardiac ventricular action potential. i.
  simulations of ionic currents and concentration changes.
\newblock {\em Circulation research}, 74:1071--96, 6 1994.

\bibitem{Ohara2011}
Thomas O'Hara, László Virág, András Varró, and Yoram Rudy.
\newblock Simulation of the undiseased human cardiac ventricular action
  potential: Model formulation and experimental validation.
\newblock {\em PLoS Computational Biology}, 7:e1002061, 5 2011.

\bibitem{Pouranbarani2019}
Elnaz Pouranbarani, Rodrigo~Weber dos Santos, and Anders Nygren.
\newblock A robust multi-objective optimization framework to capture both
  cellular and intercellular properties in cardiac cellular model tuning:
  Analyzing different regions of membrane resistance profile in parameter
  fitting.
\newblock {\em PLoS ONE}, 14, 11 2019.

\bibitem{rackauckas2020}
Christopher Rackauckas, Yingbo Ma, Julius Martensen, Collin Warner, Kirill
  Zubov, Rohit Supekar, Dominic Skinner, and Ali Ramadhan.
\newblock Universal differential equations for scientific machine learning.
\newblock {\em arXiv preprint arXiv:2001.04385}, 2020.

\bibitem{rackauckas2017}
Christopher Rackauckas and Qing Nie.
\newblock Differentialequations.jl--a performant and feature-rich ecosystem for
  solving differential equations in julia.
\newblock {\em Journal of Open Research Software}, 5(1), 2017.

\bibitem{Roscher2020}
R.~{Roscher}, B.~{Bohn}, M.~F. {Duarte}, and J.~{Garcke}.
\newblock Explainable machine learning for scientific insights and discoveries.
\newblock {\em IEEE Access}, 8:42200--42216, 2020.

\bibitem{Rush1978}
Stanley Rush and Hugh Larsen.
\newblock {A Practical Algorithm for Solving Dynamic Membrane Equations}.
\newblock {\em IEEE Transactions on Biomedical Engineering},
  BME-25(4):389--392, jul 1978.

\bibitem{Smirnov2020}
Dmitrii Smirnov, Andrey Pikunov, Roman Syunyaev, Ruslan Deviatiiarov, Oleg
  Gusev, Kedar Aras, Anna Gams, Aaron Koppel, and Igor~R. Efimov.
\newblock Genetic algorithm-based personalized models of human cardiac action
  potential.
\newblock {\em PLoS ONE}, 15, 5 2020.

\bibitem{tenTusscher2004}
K.~H.W.J.~Ten Tusscher, D.~Noble, P.~J. Noble, and A.~V. Panfilov.
\newblock A model for human ventricular tissue.
\newblock {\em American Journal of Physiology - Heart and Circulatory
  Physiology}, 286, 2004.

\end{thebibliography}

\end{document}